\documentclass[10pt,twocolumn,letterpaper]{article}

\usepackage{wacv}              

\usepackage{tikz}
\usetikzlibrary{arrows.meta,positioning,calc}
\usepackage{multirow}

\definecolor{wacvblue}{rgb}{0.21,0.49,0.74}
\usepackage[pagebackref,breaklinks,colorlinks,allcolors=wacvblue]{hyperref}

\def\wacvPaperID{*****} 
\def\confName{WACV}
\def\confYear{2027}

\title{Does the Action Matter? Probing Action-Conditioned Semantic Forecasting as a World Model for Autonomous Driving}

\author{Binay Kumar Singh \qquad Niels Da Vitoria Lobo \\ 
University of Central Florida, USA\\
{\tt\small \{binay.singh, Niels.DaVitoriaLobo\}@ucf.edu}}

\begin{document}
\maketitle
\begin{abstract}
World models are increasingly used in autonomous driving to predict how a scene
evolves under the ego vehicle's actions, supporting counterfactual reasoning and
model-based planning.
However, existing evaluations focus on prediction fidelity and planning
performance, and rarely verify that the predictions depend on the conditioning
action.
In this paper, we study action dependence in action-conditioned semantic
forecasting, which predicts future semantic segmentation from past segmentations
and the ego action.
We compare a forecaster with feature-wise and input-level action conditioning
against an identical model without action input, and measure action dependence
with test-time action interventions scored by standard segmentation metrics.
The evaluation is validated in a synthetic environment with known
action-dependent dynamics and applied to Cityscapes and BDD100K, both within
each dataset and across the two vehicle platforms.
Experiments show that the executed action improves mid-term mIoU by at most 2.5
points on Cityscapes and 0.6 points on BDD100K, whereas a wrong action reduces it by
up to 23.2 and 7.0 points, below the model without action input.
On real data, feature-wise conditioning depends 1.6 times more on the action than
input-level conditioning at similar accuracy, whereas both depend on it equally in
the synthetic environment, where the history does not determine the action.
These results show that an accuracy gain indicates that a driving world model uses
its action, but not how strongly its forecasts depend on it.
Code is available at \url{https://github.com/binayksingh/does-action-matter}.
\end{abstract}
    
\section{Introduction}
\label{sec:intro}

World models have become a central component of autonomous driving research.
By predicting future scene states conditioned on ego actions, they support
counterfactual reasoning, closed-loop simulation and model-based planning.
Recent driving world models synthesise future video conditioned on ego
trajectories or high-level commands~\cite{hu2023gaia,wang2024drivedreamer,gao2024vista},
and planning-oriented approaches learn latent or occupancy dynamics jointly with
a driving policy~\cite{hu2022mile,zheng2024occworld}.

Existing evaluations of these models focus on the fidelity of the generated
futures or on downstream planning performance, and rarely verify that the
predictions depend on the conditioning action.
Over short horizons, however, ego motion is strongly autocorrelated, so the
observed history already encodes much of the information carried by the future
action.
A model trained on logged driving data can therefore attain high forecasting
accuracy while making little use of its action input, a failure mode analogous to
the inertia and copycat problems reported for imitation-learned driving
policies~\cite{codevilla2019exploring,wen2020fighting,dehaan2019causal}.
An accuracy gain over a model without action input shows that the action is used,
but not how strongly the forecasts depend on it, since on logged data the executed
action and the action implied by the history largely coincide.
How strongly action-conditioned forecasters trained on real driving data depend on
the ego action, and whether this dependence holds across vehicle platforms,
therefore remains largely unexplored.

To address these questions, we study action dependence in semantic
forecasting~\cite{luc2017predicting}, which predicts future semantic
segmentation from past observations and has an established evaluation protocol
on Cityscapes~\cite{cordts2016cityscapes}.
Conditioning this task on future vehicle odometry has been proposed
before~\cite{bhattacharyya2019bayesian}, but whether the forecaster uses the
odometry was not examined.
We condition on the ego action, represented by speed and yaw rate, and
compare a U-Net forecaster with feature-wise linear modulation
(FiLM)~\cite{perez2018film}, input-level conditioning, and no action input.
The rationale is that semantic forecasting has a compact output space and
standard metrics, so each variant can be trained with several seeds at modest
cost and its use of the action can be measured without new evaluation criteria.
To assess whether the findings transfer across vehicle platforms, we define an
equivalent protocol on BDD100K~\cite{yu2020bdd100k}, which differs from
Cityscapes in camera, ego-motion sensing and driving environment.

To measure action dependence, we evaluate each trained model under test-time
action interventions, in which the executed future action is replaced by the
action of a sample from another sequence, by zero yaw rate, by a stop command, or
by the training-set mean action.
All conditions are scored with the standard mIoU and MO-mIoU
metrics~\cite{luc2017predicting}.
Before applying this evaluation to real driving data, we validate it in a
synthetic driving environment whose dynamics depend on the action by
construction.

Our main contributions are summarized as follows:
\begin{itemize}
  \item We present an action-conditioned semantic forecasting framework that
        compares FiLM, input-level and no action conditioning within a single
        U-Net forecaster under identical training.
  \item We develop a test-time intervention protocol that replaces the executed
        future action with shuffled, zero-yaw, stop and mean actions, and
        measures action dependence with established segmentation metrics only.
  \item We validate the interventions in a synthetic driving environment with
        action-dependent dynamics and a negative control trained on shuffled
        actions, which shows whether a model uses or ignores its action input.
  \item We provide a cross-platform evaluation on Cityscapes and BDD100K with
        GPS-derived ego actions, both within each dataset and across the two
        platforms.
\end{itemize}

\section{Related Work}
\label{sec:related}

\paragraph{World models and action-conditioned prediction.}
Predicting future observations conditioned on actions originates in
action-conditional video prediction for games~\cite{oh2015action} and robot
manipulation~\cite{finn2016unsupervised}, and in world models that learn
behaviour inside a learned simulator~\cite{ha2018world,hafner2020dream}.
In driving, MILE~\cite{hu2022mile} learns a latent world model jointly with a
driving policy, OccWorld~\cite{zheng2024occworld} forecasts 3D occupancy together
with ego motion, and GAIA-1~\cite{hu2023gaia}, DriveDreamer~\cite{wang2024drivedreamer}
and Vista~\cite{gao2024vista} synthesise future video conditioned on actions,
trajectories or commands.
A common theme across these works is that evaluation targets the fidelity of the
predicted futures or the performance of a downstream planner.
In contrast, we measure directly whether the prediction depends on the action,
using a model small enough to be trained with multiple seeds.

\paragraph{Semantic forecasting.}
Luc~\etal~\cite{luc2017predicting} introduced the forecasting of future semantic
segmentation on Cityscapes, with short-term (0.18\,s) and mid-term (0.54\,s)
horizons that subsequent work adopted.
Follow-up methods forecast with convolutional LSTMs~\cite{nabavi2018future},
recurrent flow-guided warping~\cite{terwilliger2019recurrent}, or in feature space
followed by a segmentation decoder~\cite{luc2018predicting,saric2021dense}.
These methods condition on past frames only.
Bhattacharyya~\etal~\cite{bhattacharyya2019bayesian} additionally condition a
Bayesian forecaster on past and future vehicle odometry, but report no
evaluation of how the predictions depend on it.
Our work extends this line of research by measuring that dependence under the
standard protocol, and by defining an equivalent protocol on a second platform.

\paragraph{Shortcut learning in driving models.}
Driving policies trained by imitation are known to exploit the ego vehicle's past
motion instead of the scene, which is referred to as the
inertia~\cite{codevilla2019exploring} and copycat~\cite{wen2020fighting} problems,
and more generally to rely on spurious correlates of the expert
action~\cite{dehaan2019causal}.
The forecasting counterpart of this failure mode is a world model that reproduces
the observed motion and treats its action input as optional.
Test-time input interventions provide a direct test for this behaviour, and we
adopt them as our primary evaluation.

\section{Method}
\label{sec:method}

\begin{figure*}[t]
  \centering
  \resizebox{\textwidth}{!}{\input{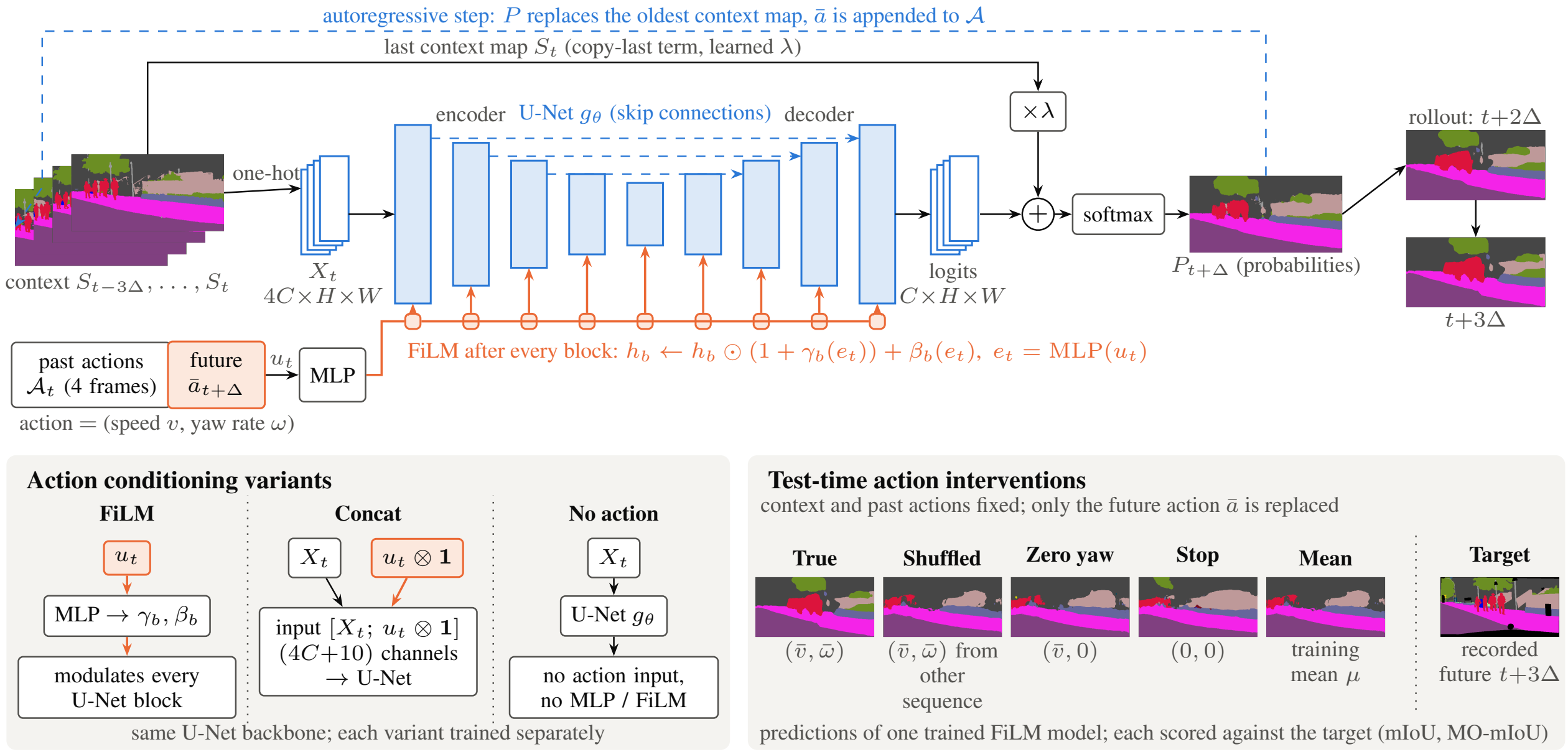}}
  \caption{Overview of action-conditioned semantic forecasting.
    A U-Net maps the four context segmentations to logits, which are added to the
    scaled last context map and normalised by a softmax; the prediction is fed back
    for the next step.
    The ego action enters through FiLM, input concatenation, or not at all (bottom
    left; the no-action model is the U-Net alone).
    At test time, the future action is replaced to measure action dependence (bottom
    right).
    The thumbnails show a Cityscapes validation scene in which the vehicle turns left
    ($\bar v = 4.6$\,m/s, $\bar\omega = 0.36$\,rad/s) and the mid-term predictions
    of a trained FiLM model; the shuffled action ($6.5$\,m/s, $0.00$\,rad/s) and the
    training-mean action ($6.3$\,m/s, $0.00$\,rad/s) both correspond to straight
    driving.}
  \label{fig:overview}
\end{figure*}

In this section, we first formalise action-conditioned semantic forecasting
(\cref{sec:task}).
We then describe the forecaster and its three action-conditioning variants
(\cref{sec:model}), the training objective (\cref{sec:objective}), and the action
interventions used to measure action dependence (\cref{sec:interventions}).

\subsection{Problem formulation}
\label{sec:task}
Let $S_t \in \{1,\dots,C\}^{H\times W}$ denote the semantic segmentation of the
frame at time $t$, with $C$ classes, and let $a_t = (v_t, \omega_t) \in \mathbb{R}^2$
denote the ego action, where $v_t$ is the speed in m/s and $\omega_t$ the yaw
rate in rad/s, positive for left turns.
Frames are sampled with a fixed step $\Delta$.
The context consists of the last four segmentations
$\mathcal{S}_t = (S_{t-3\Delta},\dots,S_t)$ and the actions at the same frames
$\mathcal{A}_t = (a_{t-3\Delta},\dots,a_t)$.
The future action over the next interval is summarised by its mean
\begin{equation}
  \bar a_{t+\Delta} = \frac{1}{\Delta}\int_{t}^{t+\Delta} a_\tau \,\mathrm{d}\tau .
  \label{eq:mean_action}
\end{equation}
Given $\mathcal{S}_t$, $\mathcal{A}_t$ and $\bar a_{t+\Delta}$, a forecaster
$f_\theta$ predicts class probabilities for the next frame,
\begin{equation}
  P_{t+\Delta} = f_\theta\big(\mathcal{S}_t,\, \mathcal{A}_t,\, \bar a_{t+\Delta}\big)
  \in [0,1]^{C\times H\times W},
  \label{eq:forecast}
\end{equation}
and the predicted segmentation is
$\hat S_{t+\Delta} = \arg\max_{c} P_{t+\Delta}[c]$.
Longer horizons are predicted autoregressively.
For step $k \ge 1$, the context is shifted by one frame, with the prediction
replacing the oldest observation and the supplied action appended to the past
actions:
\begin{equation}
  \begin{aligned}
    \mathcal{S}_{t+k\Delta} &= \big(\mathcal{S}_{t+(k-1)\Delta}[2{:}4],\; P_{t+k\Delta}\big),\\
    \mathcal{A}_{t+k\Delta} &= \big(\mathcal{A}_{t+(k-1)\Delta}[2{:}4],\; \bar a_{t+k\Delta}\big).
  \end{aligned}
  \label{eq:rollout}
\end{equation}
Following~\cite{luc2017predicting}, the short-term prediction is the first step
($t+\Delta$) and the mid-term prediction the third step ($t+3\Delta$).
During training and standard evaluation, $\bar a$ is the action executed by the
vehicle, which corresponds to the command of an ideal controller;
\cref{sec:interventions} replaces it at test time.

\subsection{Forecaster}
\label{sec:model}
\paragraph{Architecture.}
Each context segmentation is one-hot encoded (predicted probabilities are used
directly during rollout) and the four maps are stacked into an input
$X_t \in \mathbb{R}^{4C\times H\times W}$.
A U-Net~\cite{ronneberger2015unet} $g_\theta$ with four resolution levels and
$64$, $128$, $256$ and $512$ channels maps $X_t$ to $C$ logit maps.
Each block applies two $3{\times}3$ convolutions, each followed by GroupNorm and
a GELU non-linearity.
The output logits are
\begin{equation}
  \begin{aligned}
    z_{t+\Delta} &= g_\theta(X_t, u_t) + \lambda\, \mathrm{onehot}(S_t),\\
    P_{t+\Delta} &= \mathrm{softmax}\big(z_{t+\Delta}\big),
  \end{aligned}
  \label{eq:copy}
\end{equation}
where $u_t = [\mathcal{A}_t, \bar a_{t+\Delta}] \in \mathbb{R}^{10}$ is the action
vector and $\lambda$ is a learned scalar initialised to $4$.
This residual formulation ensures that the network starts from the copy-last
prediction, a strong baseline at short horizons~\cite{luc2017predicting}, and
learns only the change.
Actions are standardised with the training-set mean $\mu$ and standard deviation
$\sigma$ before entering the network.

\paragraph{Action conditioning.}
We compare three variants that share $g_\theta$ and differ only in how $u_t$
enters the network.

\emph{FiLM.} An MLP maps the action vector to a code $e_t = \mathrm{MLP}(u_t) \in \mathbb{R}^{128}$.
In every block $b$, the normalised feature map $h_b$ is modulated before the
final non-linearity~\cite{perez2018film},
\begin{equation}
  h_b \leftarrow h_b \odot \big(1 + \gamma_b(e_t)\big) + \beta_b(e_t),
  \label{eq:film}
\end{equation}
where $\gamma_b$ and $\beta_b$ are linear projections to per-channel scales and
shifts.
Both projections are initialised to zero, so that training starts from the
unconditioned network.
\emph{Concat.} The action vector is broadcast to ten constant planes and
concatenated with the input, $X_t' = [X_t;\, u_t \otimes \mathbf{1}_{H\times W}]$.
\emph{No action.} The U-Net $g_\theta$ alone, without the MLP and the FiLM
layers, which receives no action information.
The three variants have 13.07\,M (FiLM), 12.56\,M (concat) and 12.55\,M (no action)
parameters.

\subsection{Training objective}
\label{sec:objective}
The forecaster is trained with the per-pixel cross-entropy, averaged over $K{=}3$
unrolled autoregressive steps,
\begin{equation}
  \mathcal{L}(\theta) = \frac{1}{K}\sum_{k=1}^{K} \frac{1}{|\Omega_k|}
  \sum_{p\in\Omega_k} -\log P_{t+k\Delta}\big[S_{t+k\Delta}(p)\big](p),
  \label{eq:loss}
\end{equation}
where $\Omega_k$ is the set of labelled pixels at step $k$.
Unrolling exposes the model to its own predictions, as in mid-term evaluation.
For augmentation, a horizontal flip $F$ is applied jointly to the segmentations
and actions,
\begin{equation}
  (S,\; v,\; \omega) \;\mapsto\; \big(F(S),\; v,\; -\omega\big),
  \label{eq:flip}
\end{equation}
so that the flipped scene and action remain physically consistent.

\subsection{Action interventions}
\label{sec:interventions}
To test whether a forecaster uses its action input, we evaluate the same trained
model with the future actions replaced, while the observed segmentations and past
actions are kept fixed.
For sample $i$ and step $k$, the executed future action
$\bar a^{(i)}_k = (\bar v^{(i)}_k, \bar\omega^{(i)}_k)$ is replaced by
\begin{equation}
  \Phi_c\big(\bar a^{(i)}_k\big) =
  \begin{cases}
    \bar a^{(i)}_k, & c = \text{true},\\
    \bar a^{(\pi(i))}_k, & c = \text{shuffled},\\
    (\bar v^{(i)}_k,\, 0), & c = \text{zero yaw},\\
    (0,\, 0), & c = \text{stop},\\
    \mu, & c = \text{mean},
  \end{cases}
  \label{eq:interventions}
\end{equation}
where $\pi$ is a random permutation of the evaluation samples with the
constraint that $i$ and $\pi(i)$ belong to different driving sequences, since
neighbouring samples of one sequence have nearly identical actions.
The rollout of \cref{eq:rollout} is then run with $\Phi_c(\bar a)$ in place of
$\bar a$.

Each condition $c$ is scored with the standard metrics of the
task~\cite{luc2017predicting}: mIoU over all $C$ classes and MO-mIoU over the
movable-object classes.
We report the metric under each condition and its difference from the true
condition.
For the no-action model, all conditions give identical predictions by
construction; a model that has learned to ignore its action input should
likewise show differences close to zero.
In addition, the response to each condition can be checked for physical
plausibility; for example, a model that uses the action should predict a nearly
static scene under the stop condition.

\section{Experiments}
\label{sec:experiments}

We evaluate the three conditioning variants of \cref{sec:model} on Cityscapes and
BDD100K, measure their action dependence with the interventions of
\cref{sec:interventions}, and validate these interventions in a synthetic
environment.

\subsection{Datasets}
\label{sec:datasets}
\paragraph{Cityscapes.}
Cityscapes~\cite{cordts2016cityscapes} contains 30-frame video snippets recorded
at approximately 17\,Hz, with fine annotations (gtFine) for the 20th frame and
vehicle odometry for every frame.
We follow the standard forecasting protocol~\cite{luc2017predicting}: frames 8,
11, 14 and 17 are used to predict frame 20 (short term, $\Delta{=}0.18$\,s), and
frames 2, 5, 8 and 11 to predict frames 14, 17 and 20 (mid term, 0.54\,s).
Models are trained on the 2,975 training snippets and evaluated on the 500
validation snippets against gtFine at $1024{\times}2048$.
Speed and yaw rate are taken from the vehicle odometry.

\paragraph{BDD100K.}
We use the multi-object-tracking clips of BDD100K~\cite{yu2020bdd100k}, recorded
at 5\,fps and $720{\times}1280$.
A step of one frame ($\Delta{=}0.2$\,s) is the closest equivalent to the Cityscapes
step, giving short-term (0.2\,s) and mid-term (0.6\,s) horizons with the same
context structure.
One sample is drawn every 2\,s of each clip.
Speed is taken from the GPS, and the yaw rate is the negative time derivative of
the GPS course, interpolated from 1\,Hz to the frame times.
After removing clips with GPS gaps longer than 2\,s, implausible yaw rates or an
unreadable GPS file, 1,528 of the 1,600 clips remain, giving 3,658 short-term and 3,646 mid-term validation samples.
On both datasets, the distance obtained by integrating speed agrees with the GPS
distance (regression slope 1.03 and 1.008).

\paragraph{Segmentation inputs and targets.}
The BDD100K tracking clips have no dense annotations.
We therefore segment all required frames of both datasets with
SegFormer-B5~\cite{xie2021segformer} trained on Cityscapes (82.2 mIoU on
Cityscapes validation).
These segmentations are used as inputs and training targets on both datasets and
as evaluation targets on BDD100K; Cityscapes is always evaluated against gtFine.

\subsection{Implementation details}
\label{sec:impl}
Inputs are resized to $256{\times}512$ on Cityscapes and $288{\times}512$ on
BDD100K, and predictions are bilinearly upsampled to the evaluation resolution.
All models are trained from scratch with AdamW (learning rate $10^{-3}$, weight
decay $10^{-4}$), a one-cycle schedule and a batch size of 16, for 60 epochs on
Cityscapes and 12 epochs on BDD100K, with horizontal-flip augmentation
(\cref{eq:flip}).
Each variant is trained with three random seeds, and we report the mean and
standard deviation.
We compare with two baselines that use no action: copy-last, which repeats the last
observed segmentation, and flow warping, which extrapolates the optical
flow~\cite{farneback2003two} between the last two context frames to warp the last
segmentation.
On Cityscapes, we also report published results on the same
protocol~\cite{luc2017predicting,bhattacharyya2019bayesian,saric2021dense}.

\subsection{Synthetic validation environment}
\label{sec:toy}
A virtual camera with a $90^\circ$ field of view observes a horizontally wrapping
label panorama with moving cars and pedestrians; the yaw rate pans the view and
the speed zooms about the vanishing point.
The future action differs from the past action in 70\% of the samples, so that
past frames alone do not determine the next view.
In addition to the three variants, we train a negative control: a FiLM model
trained with shuffled actions, which must learn to ignore its action input.

\subsection{Results}
\label{sec:results}

\begin{table*}[t]
  \centering
  \small
  \setlength{\tabcolsep}{4pt}
  \begin{tabular}{@{}l cccc cccc@{}}
    \toprule
    & \multicolumn{4}{c}{Cityscapes (gtFine)} & \multicolumn{4}{c}{BDD100K (pseudo-labels)} \\
    \cmidrule(lr){2-5}\cmidrule(l){6-9}
    & \multicolumn{2}{c}{Short term} & \multicolumn{2}{c}{Mid term} & \multicolumn{2}{c}{Short term} & \multicolumn{2}{c}{Mid term} \\
    \cmidrule(lr){2-3}\cmidrule(lr){4-5}\cmidrule(lr){6-7}\cmidrule(l){8-9}
    Method & mIoU$\uparrow$ & MO$\uparrow$ & mIoU$\uparrow$ & MO$\uparrow$ & mIoU$\uparrow$ & MO$\uparrow$ & mIoU$\uparrow$ & MO$\uparrow$ \\
    \midrule
    Copy-last & 55.77 & 52.51 & 40.82 & 32.62 & 49.06 & 38.33 & 38.20 & 26.37 \\
    Flow warping & 66.17 & 64.82 & 46.05 & 39.79 & 52.86 & 42.43 & 39.80 & 28.50 \\
    \midrule
    S2S-dil~\cite{luc2017predicting}$^\dagger$ & 59.4 & 55.3 & 47.8 & 40.8 & -- & -- & -- & -- \\
    Bayes-WD-SL~\cite{bhattacharyya2019bayesian}$^\dagger$ & 64.8 & -- & 45.9 & -- & -- & -- & -- & -- \\
    F2MF-DN121~\cite{saric2021dense}$^\dagger$ & 69.6 & 67.7 & 57.9 & 54.6 & -- & -- & -- & -- \\
    \midrule
    No action & 70.16{\scriptsize$\pm$0.08} & 69.23{\scriptsize$\pm$0.15} & 57.02{\scriptsize$\pm$0.15} & 53.15{\scriptsize$\pm$0.28} & 54.92{\scriptsize$\pm$0.08} & 43.80{\scriptsize$\pm$0.19} & 45.04{\scriptsize$\pm$0.07} & 33.23{\scriptsize$\pm$0.21} \\
    Concat & 71.02{\scriptsize$\pm$0.06} & 69.81{\scriptsize$\pm$0.12} & 59.14{\scriptsize$\pm$0.18} & 54.60{\scriptsize$\pm$0.48} & 54.97{\scriptsize$\pm$0.07} & 43.85{\scriptsize$\pm$0.13} & 45.15{\scriptsize$\pm$0.04} & 33.14{\scriptsize$\pm$0.08} \\
    FiLM & \textbf{71.27}{\scriptsize$\pm$0.06} & \textbf{70.02}{\scriptsize$\pm$0.13} & \textbf{59.47}{\scriptsize$\pm$0.14} & \textbf{54.94}{\scriptsize$\pm$0.26} & \textbf{55.33}{\scriptsize$\pm$0.05} & \textbf{44.38}{\scriptsize$\pm$0.12} & \textbf{45.61}{\scriptsize$\pm$0.09} & \textbf{33.79}{\scriptsize$\pm$0.25} \\
    \bottomrule
  \end{tabular}
  \caption{Semantic forecasting results with the executed action (mIoU and MO-mIoU in \%; mean $\pm$ std over 3 seeds). Models are trained and evaluated on the same dataset. $^\dagger$Published results use weaker single-frame segmenters than ours and are given for reference only. Best value per column among methods run in our pipeline in bold.}
  \label{tab:main}
\end{table*}

\begin{figure}[t]
  \centering
  \includegraphics[width=\linewidth]{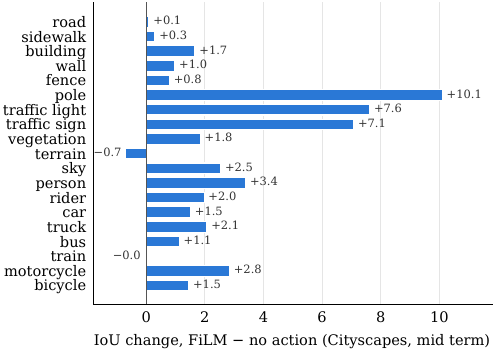}
  \caption{Per-class IoU change of FiLM over the no-action model on Cityscapes at mid
    term (executed action, mean over three seeds).}
  \label{fig:perclass}
\end{figure}

\paragraph{Main results.}
\cref{tab:main} reports forecasting accuracy with the executed action.
On Cityscapes, both action-conditioned models outperform the no-action model at
both horizons, by 1.12 (FiLM) and 0.86 (concat) mIoU at short term and by 2.46 and
2.12 mIoU at mid term.
On BDD100K, the improvement is smaller (0.40 and 0.05 at short term, 0.58 and 0.12
at mid term), and at mid term concat is below the no-action model in MO-mIoU
(33.14 vs.\ 33.23).
This indicates that the action adds accuracy mainly on Cityscapes and at the longer
horizon.
All learned forecasters outperform copy-last and flow warping; at mid term, the
no-action model alone exceeds flow warping by 10.97 mIoU on Cityscapes and 5.24 on
BDD100K, more than the action adds on either dataset.
\cref{fig:perclass} shows the per-class gains of FiLM over the no-action model on
Cityscapes at mid term.
The largest gains are for pole (10.1), traffic light (7.6) and traffic sign (7.1 IoU
points), and on BDD100K the same three classes gain most (1.51, 1.19 and 1.24).
These thin structures cover few pixels, so a small error in their predicted
position, which depends on the ego motion, removes much of their IoU.

Published results on the same protocol use different single-frame segmenters,
so the comparison is indicative only.

\begin{figure}[t]
  \centering
  \includegraphics[width=\linewidth]{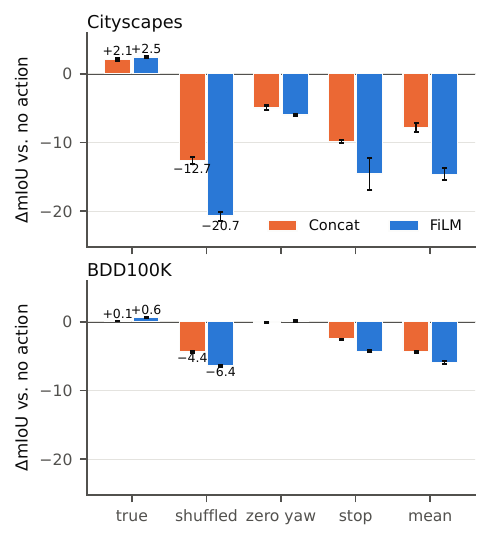}
  \caption{Change in mid-term mIoU relative to the no-action model when the future
    action is replaced at test time. Models are trained and evaluated on the same
    dataset; mean $\pm$ std over three seeds.}
  \label{fig:dependence}
\end{figure}

\begin{table}[t]
  \centering
  \footnotesize
  \setlength{\tabcolsep}{2.1pt}
  \begin{tabular}{@{}ll c cccc@{}}
    \toprule
    & & True & \multicolumn{4}{c}{Change under intervention} \\
    \cmidrule(l){4-7}
    Train / eval & Model & mIoU$\uparrow$ & Shuffled & Zero yaw & Stop & Mean \\
    \midrule
    \multirow{3}{*}{Cityscapes} & No action & 57.02{\scriptsize$\pm$0.15} & 0.00 & 0.00 & 0.00 & 0.00 \\
     & Concat & 59.14{\scriptsize$\pm$0.18} & $-$14.78 & $-$7.09 & $-$11.92 & $-$9.94 \\
     & FiLM & 59.47{\scriptsize$\pm$0.14} & $-$23.17 & $-$8.46 & $-$17.03 & $-$17.08 \\
    \midrule
    \multirow{3}{*}{BDD100K} & No action & 45.04{\scriptsize$\pm$0.07} & 0.00 & 0.00 & 0.00 & 0.00 \\
     & Concat & 45.15{\scriptsize$\pm$0.04} & $-$4.50 & $-$0.19 & $-$2.64 & $-$4.49 \\
     & FiLM & 45.61{\scriptsize$\pm$0.09} & $-$7.02 & $-$0.41 & $-$4.88 & $-$6.49 \\
    \bottomrule
  \end{tabular}
  \caption{Mid-term mIoU (\%) with the executed action and its change when the future action is replaced (shuffled: another sequence's action; zero yaw; stop; training-set mean). Changes are averaged over seeds; their standard deviation is at most 2.47.}
  \label{tab:interventions}
\end{table}

\begin{table}[t]
  \centering
  \footnotesize
  \setlength{\tabcolsep}{2.1pt}
  \begin{tabular}{@{}ll c cccc@{}}
    \toprule
    & & True & \multicolumn{4}{c}{Change under intervention} \\
    \cmidrule(l){4-7}
    Train / eval & Model & mIoU$\uparrow$ & Shuffled & Zero yaw & Stop & Mean \\
    \midrule
    \multirow{3}{*}{Cityscapes} & No action & 70.16{\scriptsize$\pm$0.08} & 0.00 & 0.00 & 0.00 & 0.00 \\
     & Concat & 71.02{\scriptsize$\pm$0.06} & $-$6.99 & $-$3.53 & $-$5.22 & $-$3.97 \\
     & FiLM & 71.27{\scriptsize$\pm$0.06} & $-$13.20 & $-$5.28 & $-$9.19 & $-$8.21 \\
    \midrule
    \multirow{3}{*}{BDD100K} & No action & 54.92{\scriptsize$\pm$0.08} & 0.00 & 0.00 & 0.00 & 0.00 \\
     & Concat & 54.97{\scriptsize$\pm$0.07} & $-$1.27 & $-$0.08 & $-$1.05 & $-$0.88 \\
     & FiLM & 55.33{\scriptsize$\pm$0.05} & $-$3.08 & $-$0.31 & $-$2.46 & $-$2.03 \\
    \bottomrule
  \end{tabular}
  \caption{Short-term mIoU (\%) with the executed action and its change under the
    interventions of \cref{tab:interventions}, averaged over seeds.}
  \label{tab:interventions_short}
\end{table}

\paragraph{Action dependence.}
\cref{tab:interventions} and \cref{fig:dependence} report the mid-term accuracy
under the action interventions; the no-action model is unaffected by construction.
First, the dependence acquired by the models exceeds the accuracy the action adds.
On Cityscapes, the shuffled action reduces mIoU by 23.17 (FiLM) and 14.78 (concat),
to 36.31 and 44.36, well below the no-action model (57.02).
Under the shuffled, stop and mean actions, both models fall below the no-action
model on both datasets.
This indicates that a wrong action harms the forecasts more than the absence of an
action.
Second, FiLM depends more on the action than concat at similar accuracy: its drop
under the shuffled action is 1.6 times larger on both datasets (23.17 vs.\ 14.78
and 7.02 vs.\ 4.50), and it is larger under every intervention.
At short term (\cref{tab:interventions_short}), all drops are smaller, since the
action affects a single forecasting step, but the ratio is larger (13.20 vs.\ 6.99
on Cityscapes and 3.08 vs.\ 1.27 on BDD100K, \ie 1.89 and 2.43), where the history
determines more of the action, and close to one in the synthetic
environment (31.57 vs.\ 30.02, \cref{tab:synthetic}), where it does not.
The difference between the mechanisms therefore appears when the action is partly
redundant with the history.
Third, all drops are smaller on BDD100K; in particular, zero yaw changes mIoU by at
most 0.41, whereas it reduces Cityscapes mIoU by 7.09 and 8.46.
This is consistent with the BDD100K actions: 90.1\% of the validation samples have a
future yaw rate below 0.05\,rad/s, compared with 81.3\% on Cityscapes.

\begin{table}[t]
  \centering
  \footnotesize
  \setlength{\tabcolsep}{4pt}
  \begin{tabular}{@{}ll cc cc@{}}
    \toprule
    & & \multicolumn{2}{c}{Eval: Cityscapes} & \multicolumn{2}{c}{Eval: BDD100K} \\
    \cmidrule(lr){3-4}\cmidrule(l){5-6}
    Train & Model & True & Shuffled & True & Shuffled \\
    \midrule
    \multirow{3}{*}{Cityscapes} & No action & 57.02{\scriptsize$\pm$0.15} & 0.00 & 40.59{\scriptsize$\pm$0.18} & 0.00 \\
     & Concat & 59.14{\scriptsize$\pm$0.18} & $-$14.78 & 42.11{\scriptsize$\pm$0.18} & $-$7.94 \\
     & FiLM & 59.47{\scriptsize$\pm$0.14} & $-$23.17 & 41.36{\scriptsize$\pm$0.09} & $-$13.42 \\
    \midrule
    \multirow{3}{*}{BDD100K} & No action & 50.36{\scriptsize$\pm$0.17} & 0.00 & 45.04{\scriptsize$\pm$0.07} & 0.00 \\
     & Concat & 50.40{\scriptsize$\pm$0.38} & $-$4.61 & 45.15{\scriptsize$\pm$0.04} & $-$4.50 \\
     & FiLM & 51.65{\scriptsize$\pm$0.03} & $-$7.78 & 45.61{\scriptsize$\pm$0.09} & $-$7.02 \\
    \bottomrule
  \end{tabular}
  \caption{Transfer between platforms: mid-term mIoU (\%) with the executed action and its change under the shuffled action, for every training and evaluation dataset.}
  \label{tab:cross}
\end{table}

\paragraph{Cross-platform transfer.}
\cref{tab:cross} reports each model on both datasets.
All models lose accuracy on the other platform: FiLM trained on Cityscapes reaches
41.36 mIoU on BDD100K (vs.\ 45.61 when trained on BDD100K), and FiLM trained on
BDD100K reaches 51.65 on Cityscapes (vs.\ 59.47).
The action still improves the transferred forecasts over the no-action model, by up
to 1.52 mIoU (concat, Cityscapes to BDD100K) and 1.29 (FiLM, BDD100K to Cityscapes).
The dependence follows the training platform rather than the evaluation data:
FiLM trained on Cityscapes loses 13.42 under the shuffled action on BDD100K, whereas
FiLM trained on BDD100K loses 7.78 on Cityscapes.

\begin{table}[t]
  \centering
  \footnotesize
  \setlength{\tabcolsep}{2.6pt}
  \begin{tabular}{@{}l c cccc@{}}
    \toprule
    & True & \multicolumn{4}{c}{Change under intervention} \\
    \cmidrule(l){3-6}
    Model & mIoU$\uparrow$ & Shuffled & Zero yaw & Stop & Mean \\
    \midrule
    Copy-last & 55.63 & -- & -- & -- & -- \\
    Flow warping & 59.40 & -- & -- & -- & -- \\
    \midrule
    No action & 59.99 & 0.00 & 0.00 & 0.00 & 0.00 \\
    Concat & 86.44 & $-$30.02 & $-$19.42 & $-$30.62 & $-$24.75 \\
    FiLM & 87.95 & $-$31.57 & $-$20.67 & $-$32.02 & $-$26.14 \\
    FiLM (shuffled training) & 60.37 & +0.02 & $-$0.01 & +0.01 & $-$0.01 \\
    \bottomrule
  \end{tabular}
  \caption{Synthetic validation environment: mid-term mIoU (\%) with the executed action and its change under each intervention (one run per model). Baselines do not take an action.}
  \label{tab:synthetic}
\end{table}

\paragraph{Synthetic validation.}
\cref{tab:synthetic} validates the interventions in the synthetic environment,
where the future view depends on the action by construction.
FiLM exceeds the no-action model by 27.97 mIoU (87.95 vs.\ 59.99), and the shuffled
action reduces the mIoU of FiLM and concat by 31.57 and 30.02.
The no-action model and the negative control trained with shuffled actions change by
at most 0.02 under every intervention.
The interventions therefore separate models that use the action from models that
ignore it.

\paragraph{Qualitative results.}
The interventions panel of \cref{fig:overview} shows the mid-term forecasts of a
FiLM model during a left turn.
Only the forecast with the executed action moves the pedestrians toward the image
centre, where they appear in the target; under the shuffled, zero-yaw, stop and mean
actions, they remain near the left image border.

\begin{table}[t]
  \centering
  \footnotesize
  \setlength{\tabcolsep}{2.3pt}
  \begin{tabular}{@{}lcccccc@{}}
    \toprule
    Method & Params & $\Delta$Params & GFLOPs & Latency (ms)$\downarrow$ & mIoU$\uparrow$ & $\Delta_\text{shuf}$ \\
           & (M)$\downarrow$ & (M) & $\downarrow$ & 1 / 3 steps & mid & mid \\
    \midrule
    No action & \textbf{12.55} & -- & \textbf{200.2} & \textbf{19.8} / \textbf{58.6} & 57.02 & 0.00 \\
    Concat & 12.56 & +0.01 & 201.7 & 20.0 / 59.1 & 59.14 & $-$14.78 \\
    FiLM & 13.07 & +0.51 & 200.2 & 20.5 / 60.7 & \textbf{59.47} & $-$23.17 \\
    \bottomrule
  \end{tabular}
  \caption{Model size and inference cost at $256{\times}512$ (fp32, batch 1, one NVIDIA V100-PCIE-32GB). $\Delta$Params is relative to the no-action model; GFLOPs count convolutions and matrix products of one forecasting step. mIoU and $\Delta_\text{shuf}$ (change under the shuffled action) are mid-term values on Cityscapes validation.}
  \label{tab:efficiency}
\end{table}

\paragraph{Model size and inference cost.}
\cref{tab:efficiency} compares the three variants.
Concat adds 0.01\,M parameters to the no-action model and FiLM adds 0.51\,M, while
both improve mid-term mIoU by 2.1 and 2.5 points, respectively.
The gain is therefore not explained by additional capacity.
Action conditioning increases the latency of one forecasting step from 19.8\,ms to
at most 20.5\,ms, and of the three-step rollout from 58.6\,ms to 60.7\,ms.

\section{Discussion}
\label{sec:discussion}
Across both datasets, the action-conditioned forecasters acquire a dependence on the
action that far exceeds the accuracy it adds.
On Cityscapes, FiLM gains 2.46 mid-term mIoU from the executed action but loses
23.17 under a shuffled one.
The accuracy gain over the no-action model shows that the action is used; the
interventions show how strongly the forecasts depend on it.
They measure the strength of this dependence, not whether it is correct.
The executed action gives the highest accuracy in every in-domain setting, and in
\cref{fig:overview} only the executed turn moves the scene as observed, which
indicates that the dependence acts in the right direction; a stronger dependence,
however, is not desirable in itself.

The difference between FiLM and concat is a property of the data as much as of the
mechanism.
Both depend almost equally on the action in the synthetic environment, where the
history does not determine it, and the gap is largest at short term, where the
history determines it most.
FiLM therefore relies on the action to a larger extent when the action is partly
redundant with the history.

The shuffled and stop interventions can be inconsistent with the observed history,
\eg a high speed after a standstill, and lie partly outside the training
distribution, so their drops combine the response to a different action with the
effect of an implausible input.
Zero yaw and the mean action stay within the range of training actions and still
reduce Cityscapes mid-term mIoU by 7.09 to 17.08 points.
Interventions drawn from samples with a similar history would separate these effects.

Models trained on BDD100K gain less from the action and depend less on it, although
the shuffled action changes the speed more on BDD100K than on Cityscapes (7.39 vs.\
4.33\,m/s on average), so the weaker dependence is not caused by a weaker
intervention.
However, the two datasets differ in camera, driving environment, frame rate and
action source, and the yaw rate interpolated from 1\,Hz GPS has a lower temporal
resolution than vehicle odometry; our experiments do not separate these factors.
On BDD100K, the targets are SegFormer-B5 segmentations rather than human
annotations, so our BDD100K conclusions are comparative between models; on
Cityscapes, evaluated against gtFine, the same ordering of the variants holds.

Our study is limited to a single front camera, a compact forecaster, semantic rather
than photometric prediction, horizons of up to 0.6\,s and two datasets, so we do not
claim that larger driving world models acquire the same dependence, nor do we relate
it to planning performance.
Our forecasts and metrics also operate at the class level.
Object-level forecasting, for which the context among co-occurring objects has been
shown to aid detection in driving scenes~\cite{singh2026contextfusion,singh2026contrastive},
would allow the interventions to be evaluated per agent.

\section{Conclusion}
\label{sec:conclusion}
In this work, we studied how strongly action-conditioned semantic forecasters depend
on the ego action.
By replacing the future action at test time and scoring the forecasts with standard
segmentation metrics, we measured action dependence for feature-wise, input-level and
no action conditioning, after validating the interventions in a synthetic
environment.
Experiments on Cityscapes and BDD100K show that the executed action improves
mid-term mIoU by at most 2.46 points, whereas a wrong action reduces it by up to 23.17
points, below the model without action input.
On real data, feature-wise conditioning depends 1.6 times more on the action than
input-level conditioning, but not in the synthetic environment, where the history
does not determine the action.
These findings show that an accuracy gain indicates that a driving world model uses
its action but not how strongly its forecasts depend on it, so action dependence
should be measured explicitly.
In future work, we plan to apply history-consistent interventions to larger video and
occupancy world models over longer horizons and to relate action dependence to
planning performance.

{
    \small
    \bibliographystyle{ieeenat_fullname}
    \bibliography{main}
}

\end{document}